\documentclass[10pt,twocolumn]{ICCAS2024}
 
\usepackage[sort&compress, numbers]{natbib}

\usepackage{amsmath} 
\usepackage{amssymb}  
\usepackage{multirow}
\usepackage{array,booktabs}
\usepackage{diagbox}
\usepackage[ruled,linesnumbered]{algorithm2e}
\SetKwInput{KwInput}{Input}
\usepackage{textcomp}
\usepackage{float}

\usepackage{irap_SIunits}
\usepackage{irap_acronyms}
\usepackage{irap_math}
\usepackage{irap_misc}
\usepackage{balance}
\usepackage{subfig}
\usepackage{rotating}
\usepackage{hyperref}

\usepackage{soul,color}
\usepackage{lipsum}
\usepackage{xcolor}
\newcommand{\bl}[1]{{\textcolor{blue}{#1}}}

\usepackage{etoolbox}
\robustify{\subref}

\usepackage[labelsep=space]{caption}

\DeclareMathOperator*{\argmin}{argmin}

\begin{document}

\title{Quantitative 3D Map Accuracy Evaluation\\
Hardware and Algorithm for LiDAR(-Inertial) SLAM}

\author{Sanghyun Hahn${}^{1^\dagger}$, Seunghun Oh${}^{2^\dagger}$, Minwoo Jung${}^{2}$, Ayoung Kim${}^{2}$, and Sangwoo Jung${}^{2*}$}

\affils{ ${}^{1}$Department of Mechanical and Aerospace Engineering, Seoul National University, \\
Seoul, 08826, Korea (steve0221@snu.ac.kr) \\
${}^{2}$Department of Mechanical Engineering, Seoul National University, \\
Seoul, 08826, Korea (alvin0808, moonshot, ayoungk, dan0130@snu.ac.kr) {\small${}^{*}$ Corresponding author}}

\thanks{ \noindent
  $^{\dagger}$The authors contributed equally to this paper.\\
 }

\abstract{
Accuracy evaluation of a 3D pointcloud map is crucial for the development of autonomous driving systems. 
In this work, we propose a user-independent software/hardware system that can quantitatively evaluate the accuracy of a 3D pointcloud map acquired from LiDAR(-Inertial) SLAM. 
We introduce a LiDAR target that functions robustly in the outdoor environment, while remaining observable by LiDAR. 
We also propose a software algorithm that automatically extracts representative points and calculates the accuracy of the 3D pointcloud map by leveraging GPS position data.
This methodology overcomes the limitations of the manual selection method, that its result varies between users.
Furthermore, two different error metrics, relative and absolute errors, are introduced to analyze the accuracy from different perspectives. 
Our implementations are available at:
\bl{\href{https://github.com/SangwooJung98/3D\_Map\_Evaluation}{https://github.com/SangwooJung98/3D\_Map\_Evaluation}}
}



\maketitle


\section{Introduction and Related Works}
\ac{LiDAR} is one of the representative sensors that provide 3D points containing accurate information around it. 
Due to its robustness of light conditions and convenient pointcloud data generation, \ac{LiDAR} is widely exploited in \ac{SLAM} research \cite{shan2020lio, reinke2022locus, xu2022fast, jung2023asynchronous, jiao2021robust}. 
With the development of \ac{LiDAR} \ac{SLAM}, the importance of an accurate 3D pointcloud map is rising as it can be exploited in various fields such as autonomous car driving \cite{wolcott2014visual, bresson2017}, high-resolution map \cite{mandikal2019dense}, and long-term map management \cite{labbe2019rtab}.

Following the increase of 3D pointcloud map utilization, evaluating the accuracy of each map has also become important. 
The accuracy of a 3D pointcloud map generated by \ac{LiDAR} \ac{SLAM} is generally approached indirectly by the accuracy of odometry, which is another output of \ac{LiDAR} \ac{SLAM}. \cite{geiger2012we}
Traditionally, the direct calculation of the accuracy of a 3D pointcloud map is performed by placing a small object at a specific location and selecting a single point that represents the object from the 3D map by human hand. 
However, due to the characteristic of \ac{LiDAR} that the pointcloud density decreases proportional to the distance from the sensor, selection of the representative point may vary depending on the user.

To overcome the problems mentioned above, we propose a target base 3D pointcloud map accuracy measurement algorithm. 
Research on targets for \ac{LiDAR} \cite{huang2021lidartag, xie2022a4lidartag, zhang20232} has been done  while their design purpose was mostly the extrinsic calibration between \ac{LiDAR} and cameras. 
Unlike existing \ac{LiDAR} targets, the proposed target is robust in outdoor environments while remaining detectable by \ac{LiDAR}. 
Furthermore, by exploiting K-means clustering \cite{kmeans1967}, \ac{RANSAC} \cite{ransac1981}, and singular value decomposition to the target pointcloud, the target position can be extracted consistently without depending on the user. 
With the calculated target pose, we propose two different error metrics (relative and absolute errors) for measuring the accuracy of 3D pointcloud maps. 
An overview of the two error metrics is illustrated in \figref{fig:figure1}, while the major contributions of the work are as follows:


\begin{figure}[!t]
    \centering
    \includegraphics[width=\columnwidth]{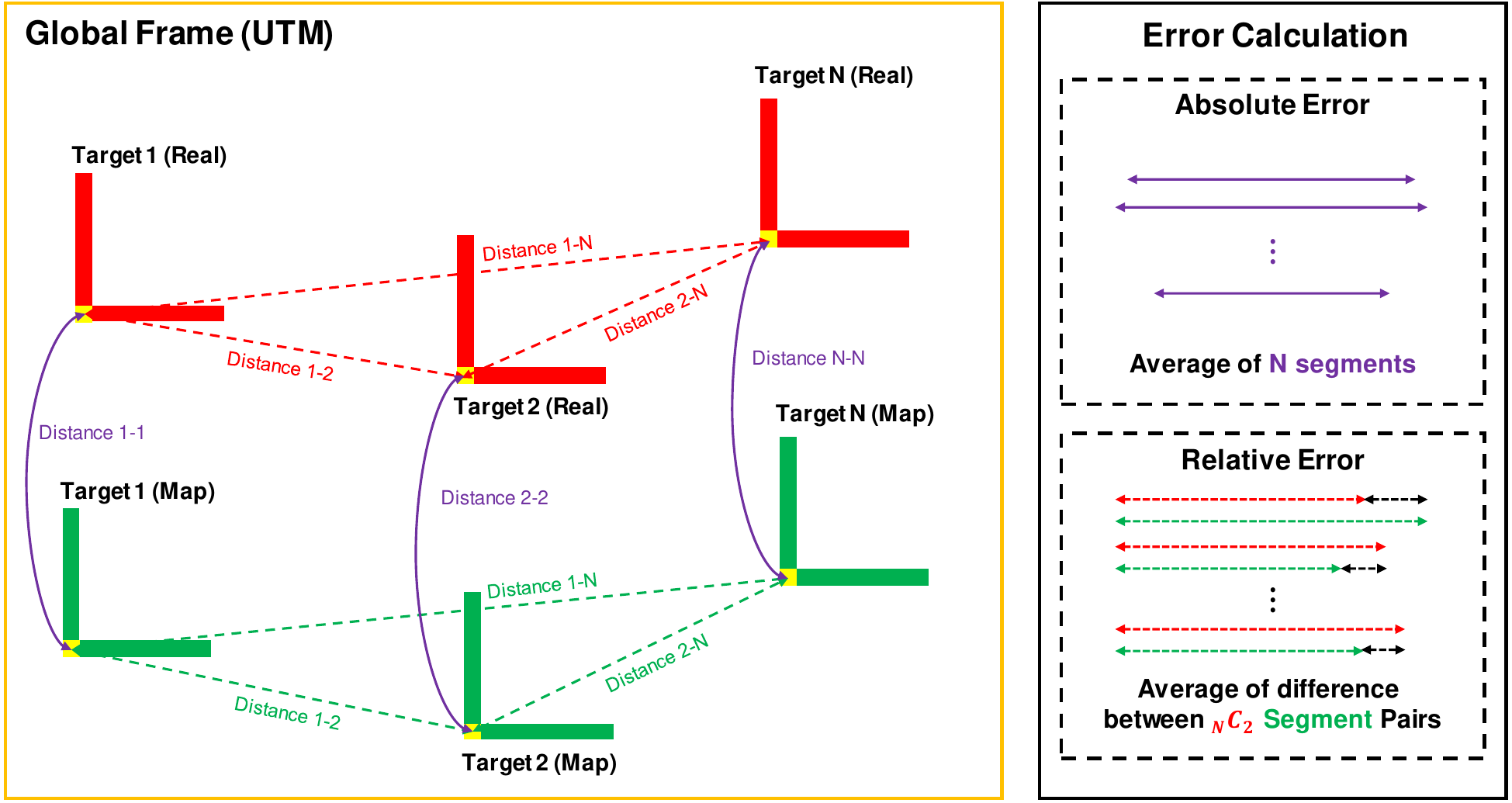}
    \caption{An overview of the absolute and relative error metrics. Absolute error is derived from the distance between $N$ pairs of corresponding target pose and ground truth pose. Relative error is derived from the distance between every two target poses.}
    \label{fig:figure1}
\end{figure}

\begin{itemize}
    \item We introduce the hardware design of the LiDAR target, which is robust in the outdoor environment, and a software algorithm that calculates the position of each LiDAR target included in the 3D pointcloud map. 
    
    \item We suggest two different error metrics (relative and absolute error) that can be exploited for analyzing the accuracy of the 3D pointcloud map from different perspectives. 

    \item We release our evaluation algorithm and hardware design to the public. 
    
\end{itemize}


\begin{figure}[!t]
    \centering
    \includegraphics[width=\columnwidth]{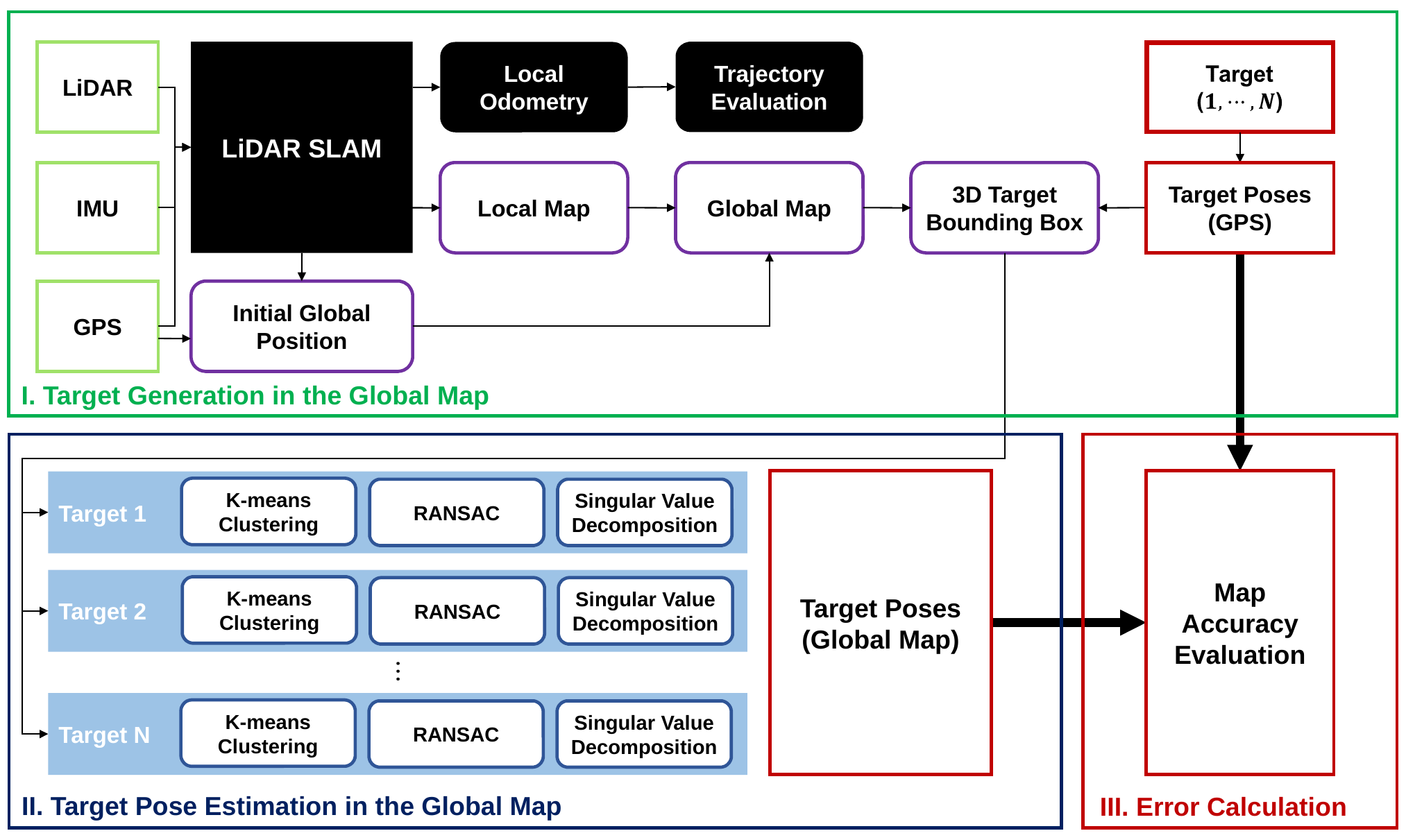}
    \caption{Pipeline of the algorithm. The target generation(green) process can be performed with any LiDAR based SLAM algorithms. Using the target pointclouds and GPS poses, target poses on the map are estimated (blue). Map accuracy is evaluated using the GPS target pose and estimated target pose (red).}
    \label{fig:pipeline}
\end{figure}

\begin{figure}[!t]
    \centering
    \subfloat[Target example 1\label{fig:target1}]{
        \includegraphics[origin=c, width=0.47\columnwidth]{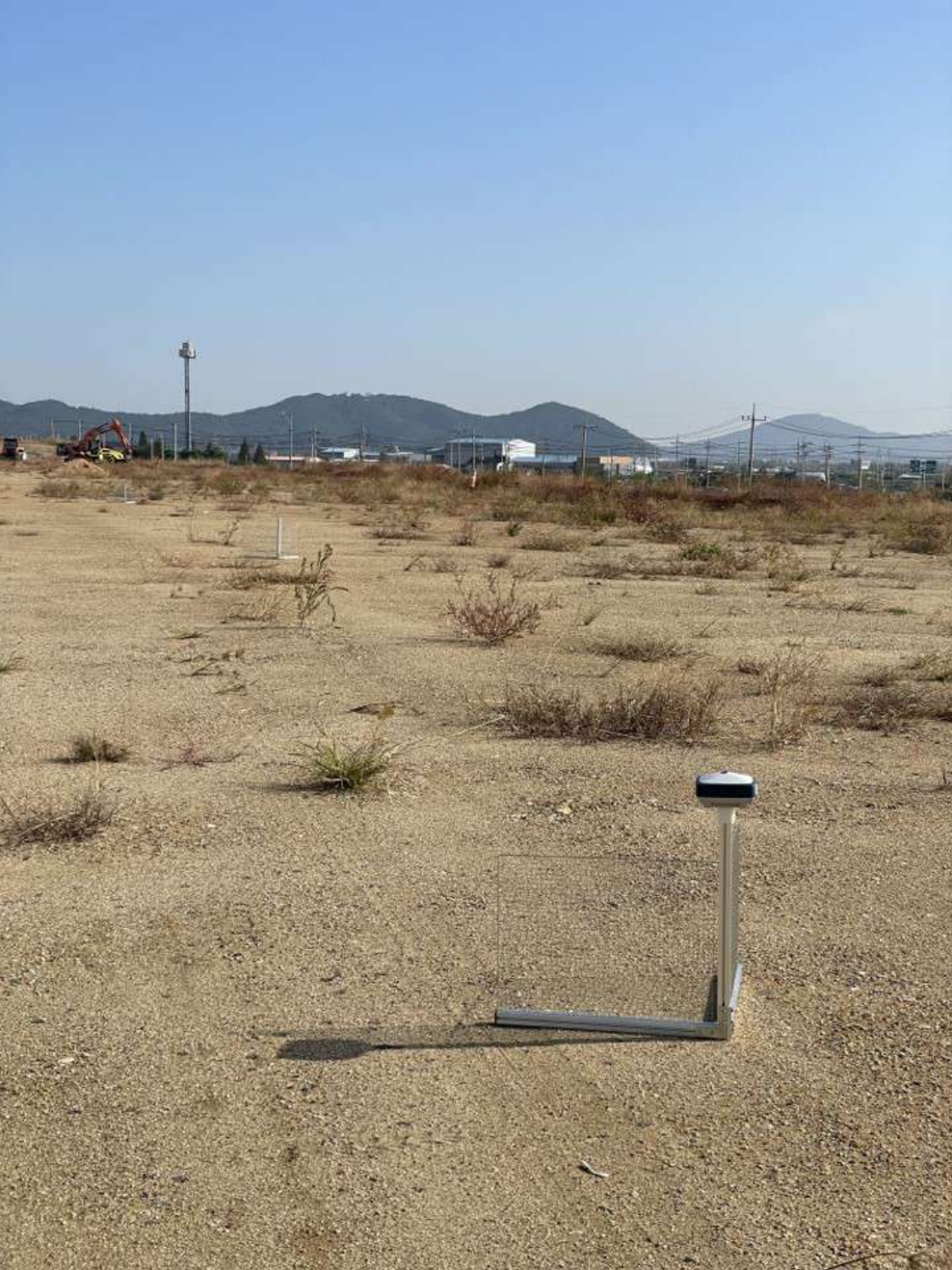}
    }
    \subfloat[Target example 2\label{fig:target2}]{
        \includegraphics[origin=c, width=0.47\columnwidth]{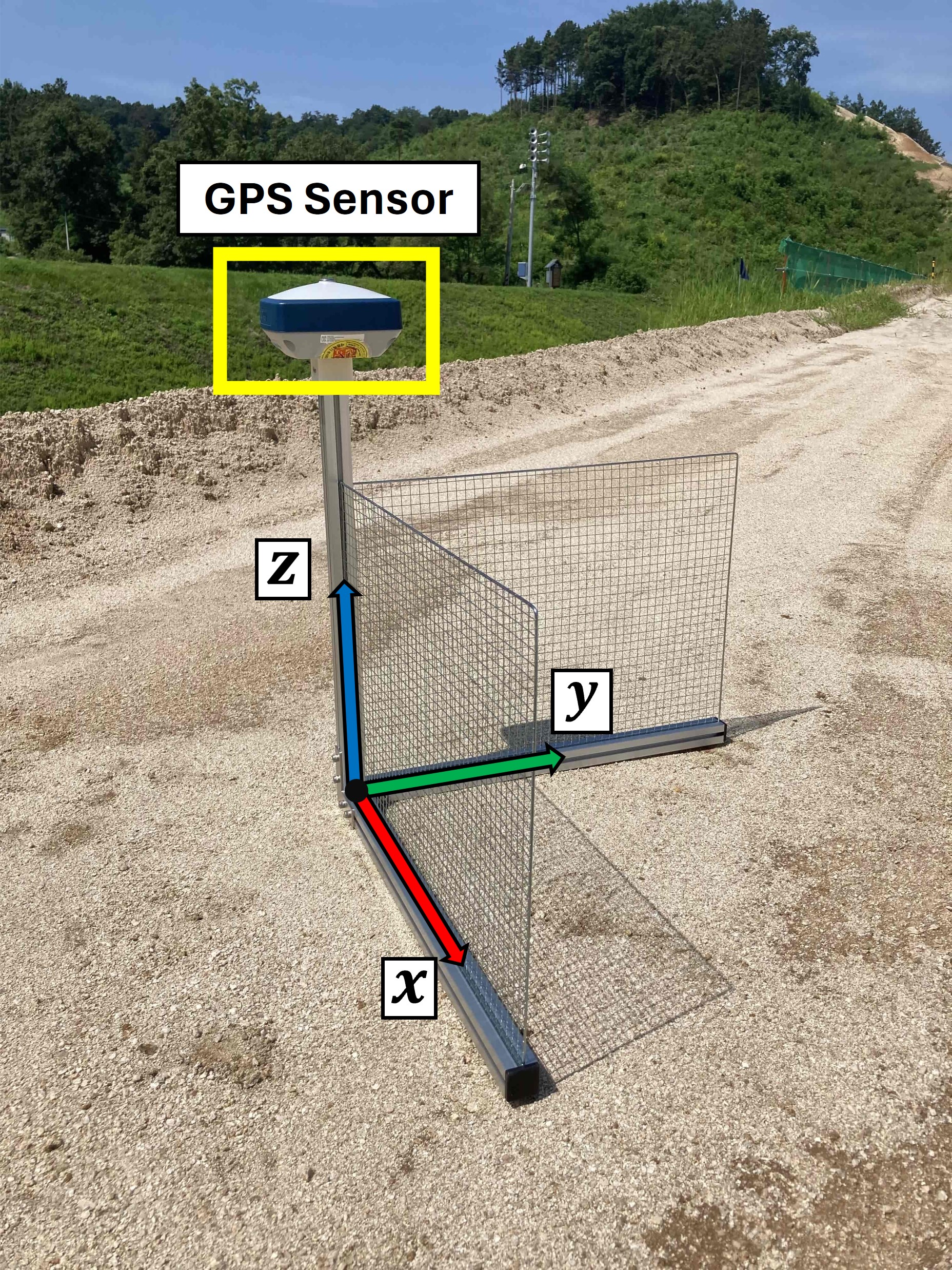}
    }
    \caption{\subref{fig:target1} and \subref{fig:target2} shows the target arrangement example. Due to the holes in the target, it is robust to wind in outdoor environments while remaining detectable by LiDAR. 
    }
    \label{fig:target}
\end{figure}

\section{Hardware Design and Target Pose Estimation}

Due to the random wind, dust, and other unpredictable factors of the outdoor environment, the target should be \textbf{1) robust to wind}, \textbf{2) GPS attachable}, and \textbf{3) easy to build}. 
Using the 30$\times$30 aluminum extrusion and cross-shaped meat grills, we built an outdoor robust \ac{LiDAR} target, as shown in \figref{fig:target}. 
The length of each edge is 0.6$\meter$, and the ground truth target pose can be acquired using the GPS sensor (Sokkia GRX3) attached to the top of it. 

\begin{figure}[!t]
    \centering
    \subfloat[Full Pointcloud\label{fig:sub1}]{
        \includegraphics[width=\columnwidth]{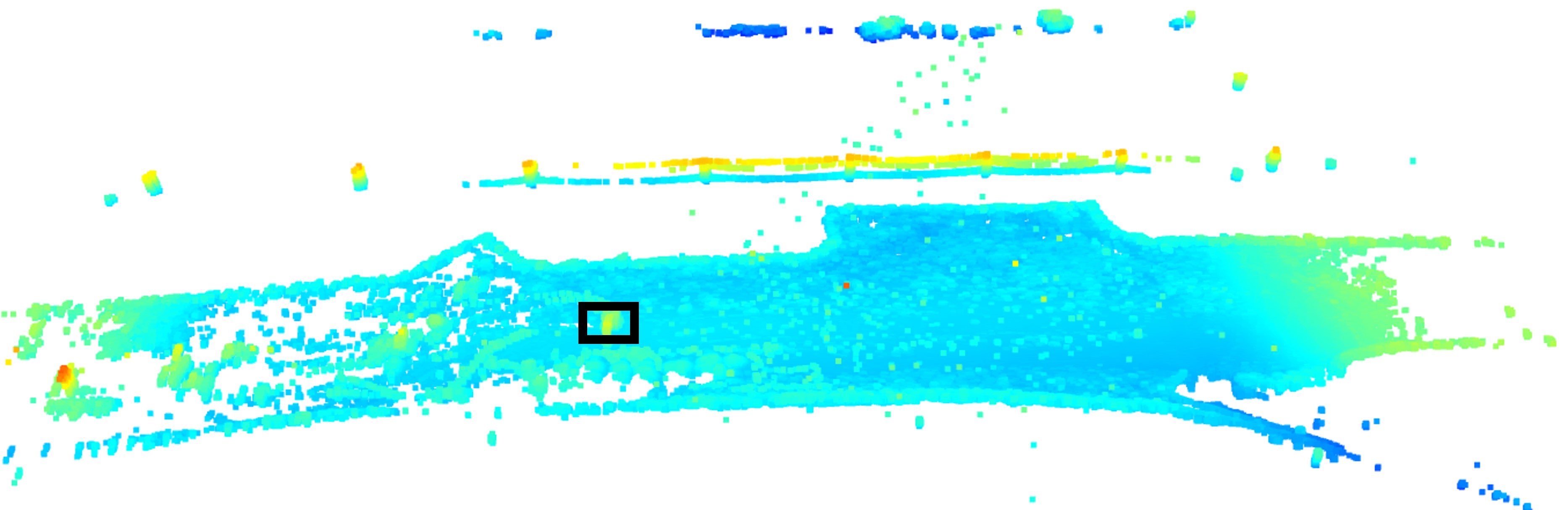}
        
    }
    \vspace{3mm}
    
    \includegraphics[width=\columnwidth]{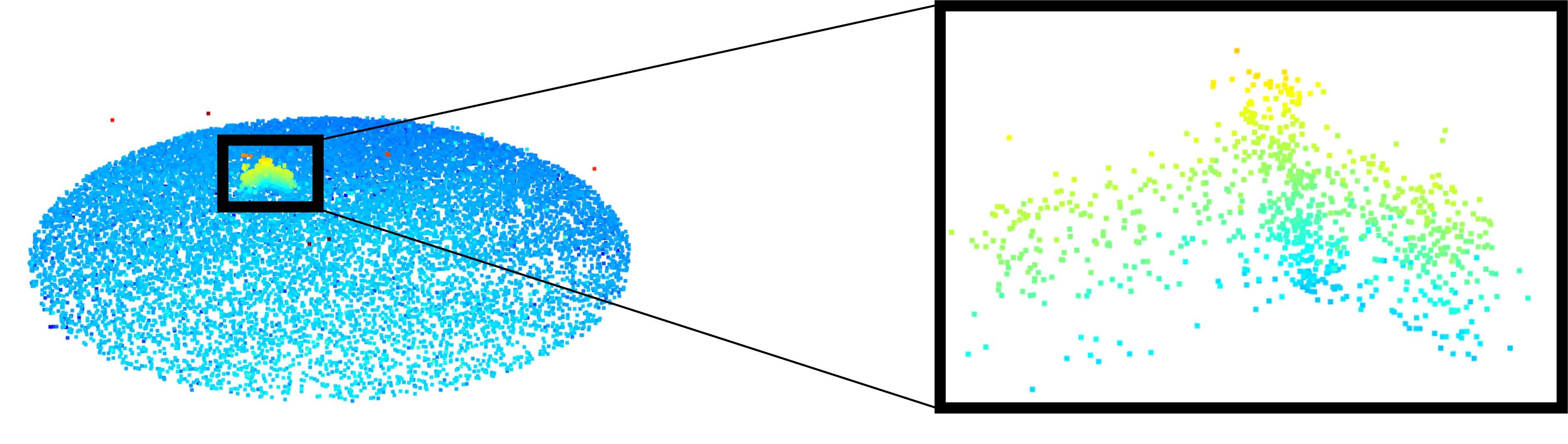}
    \vspace{-8mm}

    \subfloat[Loosely Cropped\label{fig:sub2}]{\hspace{0.42\columnwidth}} 
    \hspace{0.15\columnwidth} 
    \subfloat[Tightly Cropped\label{fig:sub3}]{\hspace{0.4\columnwidth}} 

    \caption{Visualization of the cropping process. \subref{fig:sub2} is cropped automatically from \subref{fig:sub1} based on the GPS target pose. Ground points and outliers are manually removed from \subref{fig:sub2} to obtain \subref{fig:sub3}. All black boxes indicate the same area that contains a single target.}
    \label{fig:cropping}
\end{figure}

\figref{fig:pipeline} illustrates the overall pipeline of the full algorithm. 
As an output of \ac{LiDAR} \ac{SLAM} framework, a global 3D pointcloud map is acquired. 
By cropping a sphere with a 5$\meter$ radius, centered on the GPS target pose obtained from the GPS sensor attached to the target, a pointcloud that includes the target can be cropped from the full 3D map.
This process is defined as \textbf{loose cropping}, as the resulting pointcloud includes both the target and surrounding ground points.
Alternatively, the loose cropping process can be achieved by cropping the target from the full 3D map manually without GPS.
Following loose cropping, the ground points of the loosely cropped pointcloud can be removed manually. 
This process is defined as \textbf{tight cropping}, as the resulting pointcloud mostly consists of the target itself with only a small number of outliers.
Although the tight cropping procedure is done manually, ground removal from the loosely cropped pointcloud is much easier than selecting a single point that can represent the target from the full 3D map.
As a result, the tight cropping result is more consistent among different users compared to the previous methods. 
The cropping process is displayed in \figref{fig:cropping}.

\begin{figure}[!t]
    \centering
    \begin{tabular}{@{}c@{}}
    \centering
    \subfloat[CAD modeling\label{fig:cad}]{
        \includegraphics[width=0.4\columnwidth]{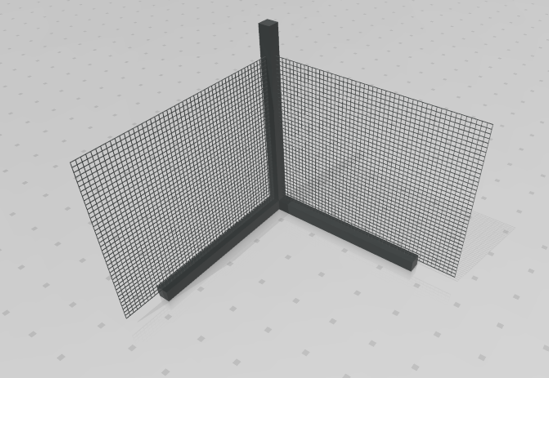}
    }
    \hspace{0.09\columnwidth}

    \subfloat[K-means Clustering\label{fig:kmean}]{
        \includegraphics[width=0.45\columnwidth]{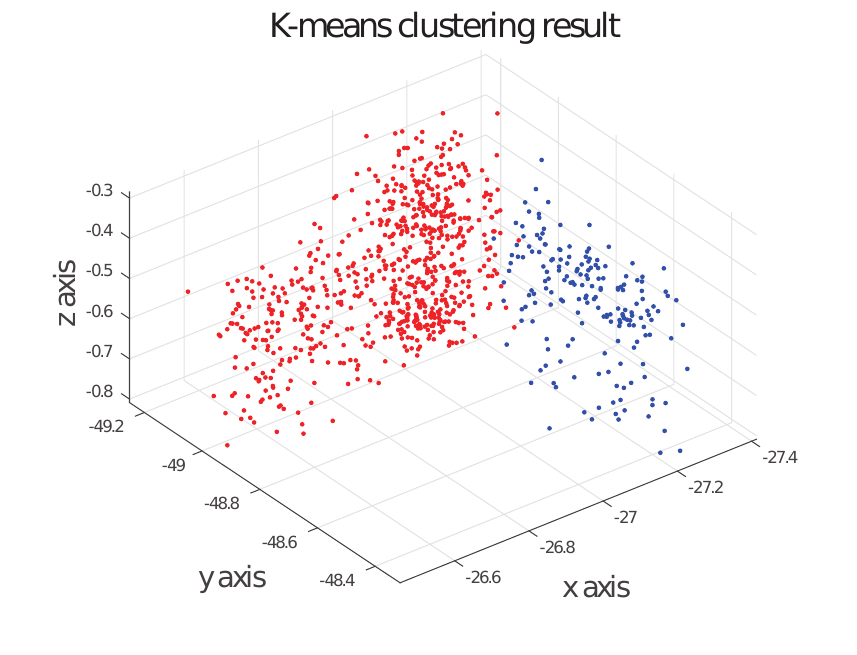}
    }
    
    \end{tabular}
    \begin{tabular}{@{}c@{}}
    \centering
    \subfloat[RANSAC\label{fig:ransac}]{
        \includegraphics[width=0.45\columnwidth]{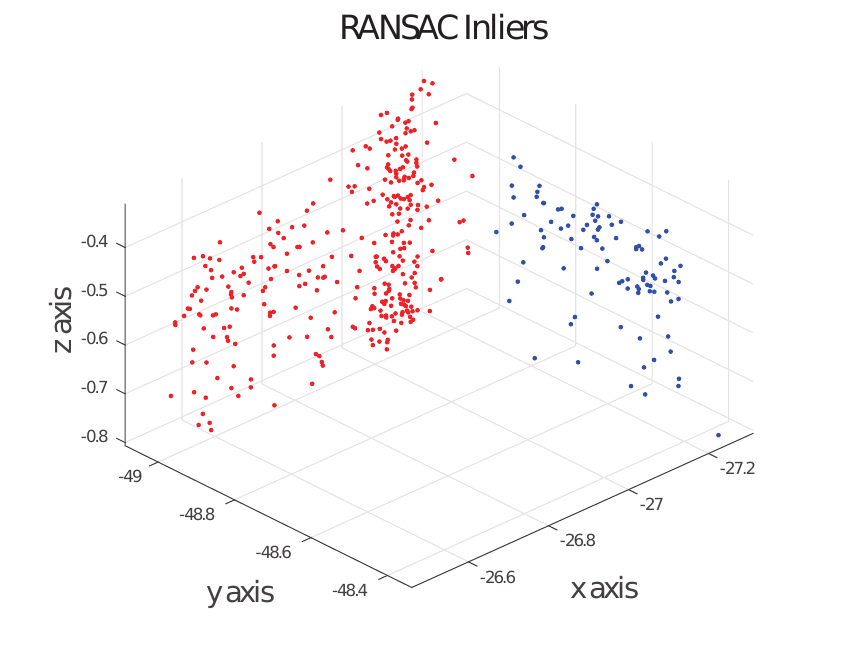}
    }
    \hspace{0.05\columnwidth}
    \subfloat[SVD\label{fig:svd}]{
        \includegraphics[width=0.45\columnwidth]{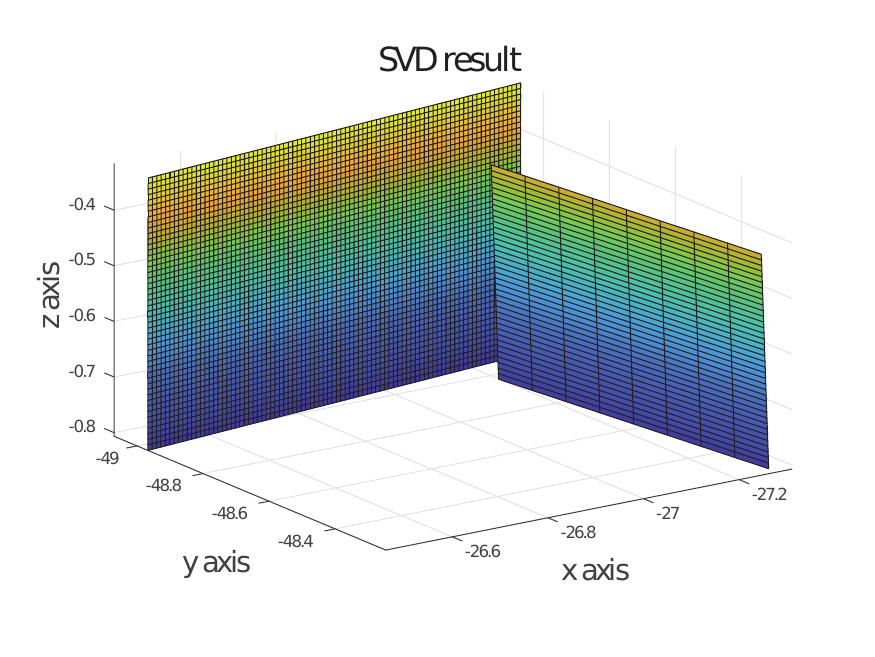}
    }
    
    \end{tabular}
    \caption{Example of target plane estimation in step by step. The red and blue points in \subref{fig:kmean} and \subref{fig:ransac} indicate the identified target planes respectively.}
    \label{fig:estimation}
\end{figure}

\begin{table*}[t]
\caption{Relative error experiment results. The relative error($E_{rel}$) and standard deviation($\sigma_{rel}$) of each sequence is written in \textbf{bold}. t1 \& t2 indicates the error in the distance between target 1 and target 2.}
\renewcommand{\arraystretch}{1.3}
\begin{center}
\resizebox{2.1\columnwidth}{!}{
\begin{tabular}{ c| c| c| c | c| c| c| c| c| c| c| c| c} \hline
\toprule
  & $E_{rel}$ (m) & $\sigma_{rel}$ (m) & t1 \& t2 & t1 \& t3 & t1 \& t4 & t1 \& t5 & t2 \& t3 & t2 \& t4 & t2 \& t5 & t3 \& t4 & t3 \& t5 & t4 \& t5 \\ 
  \midrule \hline
 Sequence 1 & \textbf{0.0754 }& \textbf{0.0413} &0.0197 &0.0826 &0.0701 &0.1271 &0.1021 &0.0902 &0.1470  &0.0140  &0.0440  &0.0569 \\ 
\midrule \hline
Sequence 2 & \textbf{0.0647}& \textbf{0.0352} & 0.0133 &0.0697 &0.0546 &0.1136 &0.0836 &0.0680  &0.1269 &0.0141 &0.0445 &0.0589 \\
\midrule \hline
Sequence 3 & \textbf{0.0499} & \textbf{0.0315}& 0.0257 &0.0748& 0.0903& 0.0932 &0.0488 &0.0650  &0.0680 & 0.0132 &0.0167 &0.0030 \\
\midrule \hline
Sequence 4 & \textbf{0.0532} & \textbf{0.0290}& 0.0056 & 0.0455 & 0.0708 & 0.0912 & 0.0521 & 0.0765 & 0.0971 & 0.0265 & 0.0461 & 0.0206 \\
\midrule \hline
Sequence 5 & \textbf{0.0633} & \textbf{0.0431} & 0.0095 & 0.0796 & 0.0988 & 0.0990  & 0.0900 &  0.1089 & 0.1096 & 0.0187 & 0.0179 & 0.0008 \\
\bottomrule \hline
\end{tabular}
}
\end{center}

\label{tab:rel_error}
\end{table*}

After extracting the tightly cropped pointcloud from the 3D pointcloud map, the calculation of its pose is achieved through the following three steps.
First, \textbf{K-means clustering} with $K=2$ is applied to the tightly cropped pointcloud, dividing it into two groups of points, $P_{1}, P_{2}$. Each cluster represents one plate of the target. 
Next, \textbf{\ac{RANSAC}} is applied to the pointcloud of each plate, $P_{1}, P_{2}$, in order to remove the outliers. 
The fitting function of \ac{RANSAC} is a plane function based on random 3 points, and the inlier threshold is 0.03$\meter$. 
After \ac{RANSAC}, inlier pointclouds that represent each plate, $P_{1}^R, P_{2}^R$ remain. 
Finally, \textbf{SVD} is applied to each inlier pointcloud, $P_{1}^R, P_{2}^R$, calculating the optimal plane function, $P_{1}^{op}, P_{2}^{op}$. 
An example of data processing at each step is displayed in \figref{fig:estimation}. 

Based on the strategy, $P_{1}^{op}, P_{2}^{op}$ should be perpendicular while each plane itself is perpendicular to the ground. 
Only plane functions that satisfy the perpendicular conditions within a threshold of 1$\degree$ proceed to the target pose calculation, while those that do not meet the threshold are redirected back to the K-means clustering process.

From the functions of $P_{1}^{op}, P_{2}^{op}$, the intersection line {$l$} is calculated as a cross of the normal vectors of $P_{1}^{op}$ and $P_{2}^{op}$.
If the target is perpendicular to the ground, the estimated target pose is ideally defined by the intersection point between {$l$} and the ground plane. 
However, inevitable errors exist in the angle between intersection line {$l$} and the ground plane.
Furthermore, the target is not exactly located at $z=0$; the target has its own height along the $z$ axis. 
Therefore, the point on $l$ corresponding to the the average height of  $P_{1}^{op}$ and $P_{2}^{op}$ is utilized as the estimated target pose.
To enhance the robustness of our algorithm, we calculated 100 sample poses for each target, averaging these to obtain the final estimated target pose. 

\begin{table}[!t]
\caption{Absolute error experiment results. The absolute error($E_{abs}$) and standard deviation($\sigma_{abs}$) of each sequence is written in \textbf{bold}.}
\renewcommand{\arraystretch}{1.3}
\begin{center}
\resizebox{1.0\columnwidth}{!}{
\begin{tabular}{ c| c| c | c| c| c| c| c}\hline
\toprule
  & $E_{abs}$(m) & $\sigma_{abs}$ (m) & target1 & target2 & target3 & target4 & target5 \\ 
  \midrule \hline
 Sequence 1 & \textbf{0.0513} & \textbf{0.0204} & 0.0523 & 0.0720 & 0.0337 & 0.0236 & 0.0752 \\ 
 \midrule \hline
 Sequence 2 & \textbf{0.0420} & \textbf{0.0211} &0.0450 & 0.0584 & 0.0282 & 0.0099 & 0.0688 \\
 \midrule \hline
 Sequence 3 & \textbf{0.0398} & \textbf{0.0098} &0.0589 & 0.0316 & 0.0361 & 0.0336 & 0.0385 \\
\midrule \hline
Sequence 4 & \textbf{0.0370} & \textbf{0.0130} &0.0403 & 0.0467 & 0.0140 & 0.0327 & 0.0511 \\
\midrule \hline
Sequence 5 & \textbf{0.0516} & \textbf{0.0110} &0.0571 & 0.0643 & 0.0313 & 0.0513 & 0.0538 \\
\bottomrule \hline
\end{tabular}
}

\end{center}
\label{tab:abs_error}
\end{table}

\begin{figure}[!t]
    \centering
    \includegraphics[width=\columnwidth]{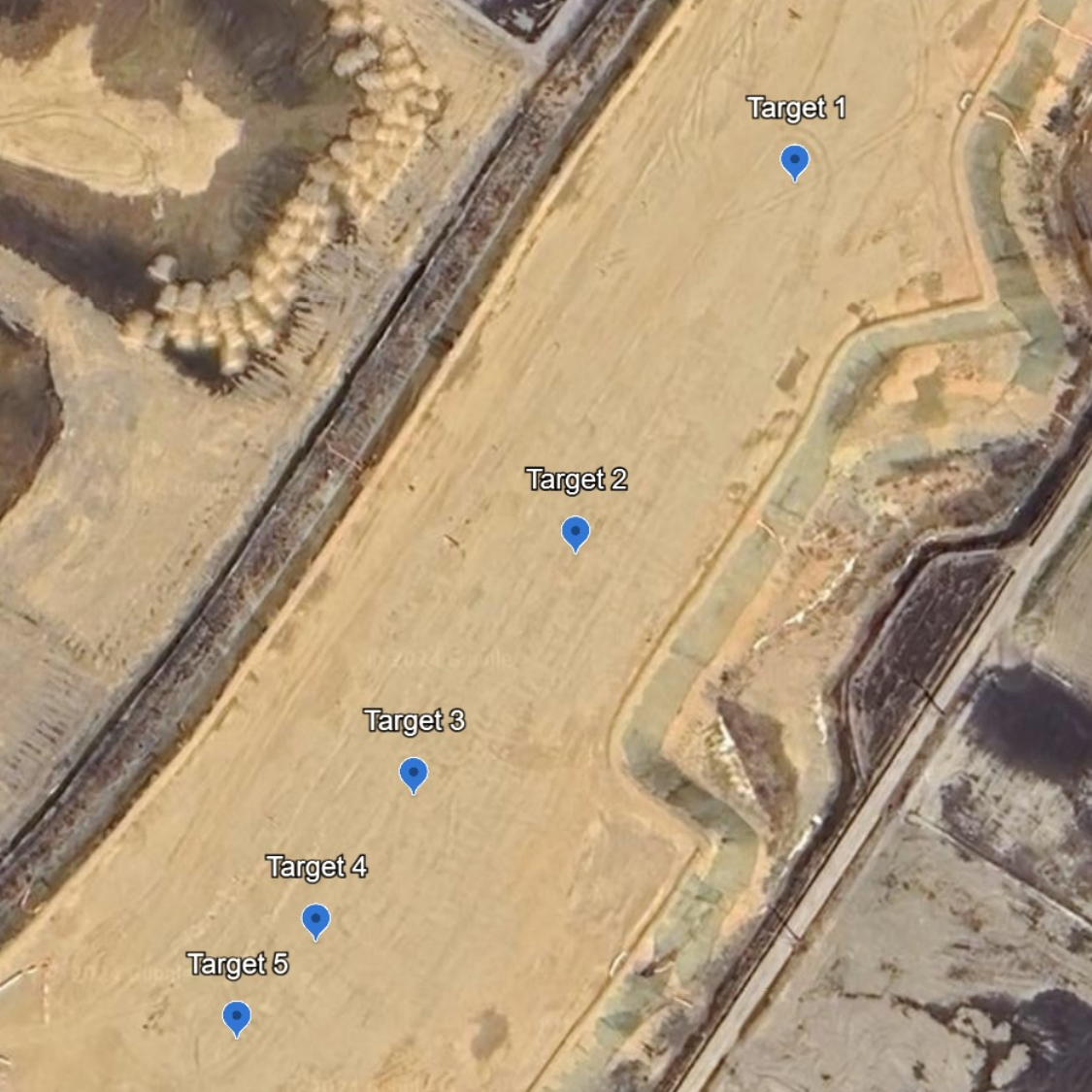}
    \caption{Satellite view example of the experimental environment and target positions. }
    \label{fig:env}
\end{figure}

\section{Relative and Absolute Error Metric}

Based on the estimated target pose and GPS-based ground truth pose, we propose two different error metrics that represent the accuracy of a 3D pointcloud map in various aspects. 
Before calculating the accuracy of the map, the estimated target pose and ground truth pose has to be aligned since the frame of the acquired 3D map may not match the frame of the ground truth pose.
We exploited a 2D image registration method based on the argmin function to minimize the sum of distances between the corresponding estimated target pose and GPS target pose.
This is achieved by translating and rotating the map frame optimally. 
The 2D transformation matrix $R$ and 2D translation vector $t$ is obtained by the following Eq. \eqref{eq:eq1}:
\begin{equation}\label{eq:eq1}
  \argmin_{R, t} \sum\limits_{i=1}^N \left\|(Rx_i + t) - \hat{x_i}\right\|_2 
\end{equation}
where N is the number of targets, $x$ and $\hat{x}$ denote the estimated target pose and GPS target pose respectively.

After registering the estimated target pose to GPS target pose, map accuracy is calculated based on two different metrics: \textbf{Relative error} and \textbf{Absolute error}. 

Relative error is calculated by averaging the difference of distance between the estimated target positions with the GPS-based distance between the same combination of targets. 
For the relative error calculation of n targets, distance error is calculated for all ${n\choose2}=\frac{n(n-1)}{2}$ combinations of targets and the average is defined as the relative error. 
In this study, 5 targets were exploited and the relative error was calculated by averaging 10 different combinations.
Since relative error is calculated based on the distance error between the targets, it focuses on evaluating the general accuracy of the map. 

Absolute error is the average of the distances between the estimated target position and its corresponding GPS-based target position.
As absolute error is calculated based on the target position error itself, the absolute error metric focuses more on the map accuracy of the local area where the targets are positioned.
Furthermore, high standard deviation of the absolute errors indicate that the accuracy of the 3D map varies across different locations of the map.

\section{Experiments and Results}

We tested the proposed algorithm based on a real-world dataset acquired from a highway construction site.
The 3D pointcloud map was generated by the \ac{LiDAR} \ac{SLAM} algorithm LIO-SAM\cite{shan2020lio}, while Velodyne VLP-32C \ac{LiDAR}, MicroStrain 3DM-GX5-25 9 DoF IMU, and NovAtel CPT7 GPS were exploited as the hardware. 
The experiment environment and hardware position are displayed in \figref{fig:env}. 

With consistent target and environment settings, we acquired 5 sequences and calculated the relative and absolute errors for each sequence. 
The experiment results are displayed in \tabref{tab:rel_error} and \tabref{tab:abs_error}.

The experiment results show that relative error and absolute error have similar tendencies following the sequences. 
Sequences with smaller relative error show small absolute error, and vice versa. 
This indicates that the general map error of the 3D map is highly related to the local map error, suggesting that one error metric includes information of the other error metric indirectly. 

Furthermore, as presented in \tabref{tab:rel_error}, there is a tendency for the relative error to increase as the distance between targets increases.
Due to this effect, relative error may show higher values for larger maps.
On the other hand, absolute error compares the position of a single target based on GPS target pose and estimated target pose, providing map-size-independent accuracy information.

For absolute error, the results displayed in \tabref{tab:abs_error} indicate that the error from targets 2, 3, and 4 tends to be lower than those from targets 1 and 5. 
This is due to the robot's trajectory which makes an ellipsoid around the targets, visiting targets 1 and 5 only once while other targets are visited twice. 

It is challenging to keep the factors such as the distance from the robot to each target and the number of visits per target constant across different sequences. 
In order to remove these effects and measure the accuracy of the 3D pointcloud map robustly, the relative error metric may be exploited.

\section{Conclusion}

In this work, we propose hardware and a software algorithm that can be exploited to measure the accuracy of a 3D pointcloud map while maintaining the robustness in outdoor environments and LiDAR sparsity. 
Furthermore, we introduce two different error metrics, Relative error and Absolute error, that represent general map accuracy and local map accuracy. 
Through the real-world experiment, we demonstrated the relationship between two error metrics and the robustness of our hardware and algorithm.
This approach offers insights into the local and global quality of the 3D pointcloud map, enhancing evaluation standards for LiDAR(-Inertial) SLAM algorithms.
To overcome the current limitations of this research, we plan to develop a fully automated software that operates without any manual intervention.


\bibliographystyle{IEEEtranN} 
\bibliography{string-short,root}

\end{document}